\UseRawInputEncoding
\documentclass{midl} % Include author names
\usepackage{mwe} % to get dummy images
\usepackage[skip=0pt]{caption}
\usepackage{wrapfig}
\usepackage{multirow}
\usepackage{graphicx}
\usepackage{lscape}
\usepackage{longtable}
\usepackage{float}
\usepackage{scrextend}
\jmlrproceedings{}{}
\jmlrpages{}
\jmlryear{}

\jmlrworkshop{}
\editors{}

\title[The statistical association between CAM and clinical domain knowledge]{Opening the black box of deep learning: Validating the statistical association between explainable artificial intelligence (XAI) and clinical domain knowledge in fundus image-based glaucoma diagnosis}

\midlauthor{\Name{Han Yuan\nametag{$^{1}$}} \Email{yuan.han@u.duke.nus.edu}\newline
\Name{Lican Kang\nametag{$^{2}$}} \Email{kanglican@whu.edu.cn}\newline
\Name{Yong Li\nametag{$^{3,4}$}} \Email{liyong@u.duke.nus.edu}\newline
\addr $^{1}$Centre for Quantitative Medicine, Duke-NUS Medical School\newline
\addr $^{2}$Cardiovascular and Metabolic Disorders Programme, Duke-NUS Medical School
\newline
\addr $^{3}$Department of Ophthalmology and Visual Sciences, Duke-NUS Medical School\newline
\addr $^{4}$Singapore Eye Research Institute, Singapore National Eye Centre\newline
}

\begin{document}

\maketitle
\begin{abstract}
While deep learning has exhibited remarkable predictive capabilities in various medical image tasks, its inherent black-box nature has hindered its widespread implementation in real-world healthcare settings. Our objective is to unveil the decision-making processes of deep learning models in the context of glaucoma classification by employing several Class Activation Map (CAM) techniques to generate model focus regions and comparing them with clinical domain knowledge of the anatomical area (optic cup, optic disk, and blood vessels). Four deep neural networks, including VGG-11, ResNet-18, DeiT-Tiny, and Swin Transformer-Tiny, were developed using binary diagnostic labels of glaucoma and five CAM methods (Grad-CAM, XGrad-CAM, Score-CAM, Eigen-CAM, and Layer-CAM) were employed to highlight the model focus area. We applied the paired-sample t-test to compare the percentage of anatomies in the model focus area to the proportion of anatomies in the entire image. After that, Pearson's and Spearman's correlation tests were implemented to examine the relationship between model predictive ability and the percentage of anatomical structures in the model focus area. On five public glaucoma datasets, all deep learning models consistently displayed statistically significantly higher percentages of anatomical structures in the focus area than the proportions of anatomical structures in the entire image. Also, we validated the positive relationship between the percentage of anatomical structures in the focus area and model predictive performance. Our study provides evidence of the convergence of decision logic between deep neural networks and human clinicians through rigorous statistical tests. We anticipate that it can help alleviate clinicians' concerns regarding the trustworthiness of deep learning in healthcare. For reproducibility, the code and dataset have been released at GitHub\footnote{https://github.com/Han-Yuan-Med/the-statistical-association-between-cam-and-domain-knowledge\label{note1}}.
\end{abstract}

\newpage
\section{Introduction}
In the last decade, deep learning has reshaped various ophthalmological disease diagnoses \cite{yong2024}. Although deep neural networks feature high-fidelity accuracy on retrospective classification, localization, and segmentation tasks, clinicians still cannot fully trust their decisions on prospective medical practice due to their black-box characteristics \cite{rudin2019stop}. To open the black box of deep learning and address the interpretability issue, various methods have been proposed to explain model inference logic on different healthcare data \cite{2024automl}. In the realm of medical image analysis, a commonly used set of methods, known as the Class Activation Map (CAM) family, is developed to map the final diagnostic decision back onto the input images by highlighting the important pixels (regions) within the images. Such an assisted heatmap will be provided to clinicians to conduct further evaluation on whether the model focuses on the clinically relevant region of interest (ROI) \cite{cam}. If a deep learning model consistently performs well on a specialized task and puts its attention on the lesion area causing the disease, clinicians' concern about the model reliability would be alleviated \cite{2024mldeployment}.

Glaucoma is a neuro-degenerative disease caused by the increase of intra-ocular pressure and will progress into complete blindness without early medical intervention \cite{varma1992expert,liang1997supernormal,garway1998vertical,morgan2012accuracy}. Conventional clinical diagnosis was based on the clinician’s manual evaluation of optic cup, optic disk, blood vessels, intraocular pressure, and visual field \cite{varma1992expert,liang1997supernormal,garway1998vertical,morgan2012accuracy,xue2022multi,morano2021simultaneous}. Recently, multiple deep learning-based systems were proposed to automate this process \cite{thompson2020review,mirzania2021applications}. For example, Shinde introduced a comprehensive four-stage pipeline for glaucoma classification, including the delineation of the optic disk area, the segmentation of the optic cup and disk, the extraction of relevant clinical features, and finally, the classification of glaucoma based on these previously extracted features \cite{shinde2021glaucoma}. However, prior research efforts in this field have primarily focused on enhancing the diagnostic accuracy of glaucoma \cite{mitra2018region,zhao2023joint,liu2022ecsd,zhao2019direct}. While some studies have offered qualitative visualizations illustrating that the decision logic of deep neural networks aligns with clinical domain knowledge concerning the optic cup, optic disk, and blood vessels \cite{akter2022glaucoma,thakoor2019enhancing,li2019large,liao2019clinical}, there remains a need for quantitative and systematic evaluations on model decision logic to further bolster the confidence of ophthalmologists in the deployment of deep neural networks.

To address the quantification issue above, we implemented a paired-sample t-test to measure the association between model focus region and domain knowledge-based anatomical area in the context of glaucoma diagnosis. Compared with the clinically irrelevant area in the input image, the model statistically significantly concentrated more on the anatomical area (optic cup, optic disk, blood vessels) where clinicians make their diagnosis, showing the convergence of data-driven solutions and human knowledge-based strategies \cite{varma1992expert,liang1997supernormal,garway1998vertical,2024consxai}. We further demonstrated that the model diagnostic performance was positively correlated with its attention ratio on the important lesion area, indicating that future model developers should take the clinical knowledge into account rather than treat medical image tasks as general computer vision missions using a fully end-to-end approach. Compared with the previous work on comparing model explanation and domain knowledge \cite{liao2019clinical,guide}, we performed rigorous statistical tests on both convolutional neural networks (CNN) and transformers. In summary, we presented that an image-level diagnostic annotations-based model correctly summarized reasonable medical knowledge towards clinicians, underscoring the potentiality of deep learning in distilling latent knowledge in the future machine intelligence-based machine \cite{2024CDM}.

\section{Materials and Methods}
\subsection{Dataset}
We conducted a comprehensive investigation into the association between deep learning inference logic and clinical domain knowledge by utilizing five public datasets: ORIGA \cite{Origa}, HRF \cite{HRF}, LES-AV \cite{LES-AV}, Drishti-GS \cite{Drishti}, and FIVES \cite{Fives}. These datasets encompassed a total of 650, 30, 22, 101, and 300 fundus images, respectively. Table \ref{tab:dataset-overview} provides an overview of the used datasets. All fundus images from the five datasets were resized to the resolution of 224$\times$224 pixels to meet the requirements of most pre-trained deep learning backbones. In addition to binary diagnostic labels indicating the presence or absence of glaucoma, ORIGA and Drishti-GS were enriched with pixel-level annotations for anatomical structures of the optic cup and disk while HRF, LES-AV, and FIVES contained delineations of blood vessels. ORIGA, HRF, and LES-AV were used to develop glaucoma classifiers, conduct model explanations, and validate the association between model focus area and domain knowledge-based anatomical structures. Drishti-GS and FIVES were introduced as external evaluation of model classification and CAM explanation. To maximize the utility of the limited samples with pixel-level annotations of blood vessels, we implemented double cross-validation in model development and internal evaluation. This involved initially dividing the dataset into three non-overlapped subsets, followed by the allocation of training, validation, and test datasets to each of these subsets. Ultimately, this process generated $A_{3}^{3}$: 6 distinct scenarios to comprehensively evaluate various glaucoma classifiers and the model focus areas generated by different CAM methods \cite{burzykowski2023validation}. Table \ref{tab:data-split} shows the concrete details of the double cross-validation. For the external evaluation on Drishti-GS and FIVES, we randomly divided the dataset into non-overlapping validation and test sets with a 50:50 ratio.

\subsection{Deep learning-based glaucoma classification}
We first give necessary notations to facilitate the downstream elaboration on classifier training and explanation of model decision logic in terms of focus area. For the glaucoma classification task, we denote the training, validation, and test datasets as $D^{train}$, $D^{val}$, and $D^{test}$, respectively. Take the training dataset $D^{train}$ as an example, each fundus image $I_i^{train}$ is paired with image-level diagnostic label $Y_i^{train}$ and pixel-level annotation $A_i^{train}$ of either optic cup and disk or blood vessels. $I_i^{train}$ designates a two-dimensional image with a width of $W_0$ and a height of $H_0$ and $p_{w,h}(I_i^{train})$ denotes a pixel in $I_i^{train}$ whose coordinate of width and height is $(w,h)$. For the generation of labels, clinicians screened each $I_i^{train}$ and assigned the binary diagnostic label of $Y_i^{train}$ in which 1 stands for glaucoma. Additionally, clinicians delineated the anatomical structure, $A_i^{train}$, marking the pixels belonging to this specific anatomical structure with 1. The glaucoma classifier is trained based on $I_i^{train}$ and $Y_i^{train}$, validated using $I_i^{val}$ and $Y_i^{val}$, and tested by $I_i^{test}$ and $Y_i^{test}$. After the model development, $A_i^{test}$ is adopted to quantify the association between clinical domain knowledge of anatomical structures and the model decision logic in terms of the focus area.

Specifically, the training process of deep neural networks is to find a set of parameters for a pre-defined architecture that minimizes the difference between model predictions and ground truth labels in the training set. Formally, with the training dataset $D^{train}$, we aim to optimize a deep learning model $f_{\theta}$ parameterized by $\theta$. Taking the input of $I_i^{train}$, $f_{\theta}$ outputs $f_{\theta}(I_i^{train})$ and the optimization target is to minimize the loss function $l$ between $f_{\theta}(I_i^{train})$ and sample labels $Y_i^{train}$ across all samples in $D^{train}$. To facilitate fine-grained training of $f_{\theta}$, the validation dataset $D^{val}$ is applied to schedule the learning rate decay: If $l(f_{\theta}(I_i^{val}),Y_i^{val})$ has not decreased for a pre-defined epoch number $N_{epoch}$, the learning rate of $\theta$ will be decreased and the $\theta$ showing the best performance on $l(\theta;D^{val})$ will be saved as the optimal parameter $\theta^*$. After the model training, we calculate the binarization threshold $\tau$ based on the model's optimal performance on $D^{val}$ and then evaluate the classification performance of the trained $f_{\theta^*}$ on the unseen test dataset $D^{test}$. Various metrics are applied to quantify the model performance by comparing the model prediction $f_{\theta^*}(I_i^{test})$ and the ground truth label $Y_i^{test}$. In this study, we utilize seven common metrics for glaucoma classification evaluation, including the area under the receiver operating characteristic curve (AUROC), the area under the precision recall curve (AUPRC), accuracy, sensitivity, specificity, positive predictive value (PPV), and negative predictive value (NPV). Also, the standard error (SE) is reported based on the bootstrapping of samples in the test dataset \cite{bootstrap}. Besides the internal evaluation above, we introduce the external unseen validation set $D^{val}_{ext}$ and test set $D^{test}_{ext}$ for generalizability evaluation. Without fine-tuning, the developed model $f_{\theta^*}$ is deployed on $D^{val}_{ext}$ and $D^{test}_{ext}$ for binarization threshold determination and model performance evaluation. 

Considering the middle-scale characteristic of the used dataset, we developed glaucoma classifiers using four lightweight deep neural networks \cite{2024resource,2025foundation}. These classifiers encompass two CNN backbones: VGG (VGG-11) \cite{vgg} and ResNet (ResNet-18) \cite{he2016deep}, as well as two Transformer architectures: Vision Transformer (DeiT-Tiny) \cite{dosovitskiy2020image,touvron2021training} and Swin Transformer (Swin-Tiny) \cite{liu2021swin}. For model training, we employed Stochastic Gradient Descent (SGD) \cite{SGD} with a learning rate of 0.001, a momentum of 0.9, and a decay of 0.9 with a patience parameter of 10. Each deep learning model underwent training for 100 epochs with weighted samples \cite{imbalance} across the six-fold cross-validation scenarios mentioned above. We reported the model performance with SE on both internal and external test datasets.

\subsection{CAM explanation for deep learning classifiers}
The developed model $f_{\theta^*}$ classifies an unseen image $I_i^{test}$ from the test dataset $D^{test}$ as $f_{\theta^*}(I_i^{test})$. To explain the model decision logic behind $f_{\theta^*}(I_i^{test})$, the model focus area $R(I_i^{test})$ in $I_i^{test}$ is extracted by various CAM methods to highlight the most important pixels (regions) towards the model final decision $f_{\theta^*}(I_i^{test})$. CAM techniques calculate each pixel's importance towards $f_{\theta^*}(I_i^{test})$ and $E_{(w,h)}(I_i^{test})$ denoted the importance of a pixel with coordinate $(w,h)$ in $I_i^{test}$. We further outline the focus area $R(I_i^{test})$ consisting of the most significant pixels by selecting the pixels with top 5\% $E_{(w,h)}(I_i^{test})$.

In this study, we utilize five mainstream explanation techniques in the CAM family: Grad-CAM \cite{gradcam}, XGrad-CAM \cite{fu2020axiom}, Score-CAM \cite{wang2020score}, Eigen-CAM \cite{muhammad2020eigen}, and Layer-CAM \cite{jiang2021layercam} to generate the pixel-level importance $E_{(w,h)}(I_i^{test})$. The top 5\% important pixels are extracted to formulate the model focus area $R(I_i^{test})$. After that, we compare the percentages of anatomical structures in the model focus area $(R(I_i^{test}){\bigcap}A_i^{test})/R(I_i^{test})$ and the whole image $A_i^{test}/I_i^{test}$. Similar to the external classification evaluation, we also implement external evaluation on model explanation using $D^{test}_{ext}$. Apparently, if deep learning models allocate no additional attention to anatomical structures compared with other regions, there would be no significant difference between the percentage of anatomies in the model focus area and the entire input image. However, if the model identifies glaucoma predominantly by relying on evidence from anatomical structures, the first percentage would be significantly higher than the second.

To establish a rigorous analysis of the relationship between model decision logic (focus area) and medical domain knowledge (the optic cup, optic disk, and blood vessels), the paired-sample t-test \cite{student1908probable} is employed to compare the percentages of anatomical structures in the model focus area and the whole image. Besides, we compute both Pearson's \cite{pearson1920notes} and Spearman's correlation coefficient \cite{fieller1957tests} to ascertain whether there exists a correlation between the model's classification performance of AUROC and the proportion of anatomical areas within its focus region. A correlation coefficient larger than 0 implies a positive correlation between the two variables. For reproduction, code and dataset are publicly available at GitHub\footref{note1}.

\section{Results}
First, we quantitatively demonstrated the internal classification performance of various models on glaucoma diagnosis in Table \ref{tab:classification-int} and their external performance on Grishti-GS and FIVES in Table \ref{tab:classification-drishiti} and Table \ref{tab:classification-fives}, respectively. 
In both internal and external evaluations, VGG-11 consistently outperformed other models, achieving the highest classification performance across all metrics, with the exception of sensitivity in the external Drishti-GS dataset. ResNet-18, DeiT-Tiny, and Swin-Tiny also delivered a commendable performance in internal and external glaucoma classification, demonstrating the feasibility of downstream analysis of the models' decision logic.

Table \ref{tab:VGG-11}, \ref{tab:ResNet-18}, \ref{tab:DeiT-Tiny}, and \ref{tab:Swin-Tiny} presents the CAM explanation performance of different deep learning models on the internal test dataset and their external explanation performance on Drishti-GS and FIVES was provided in Table \ref{tab:vgg11ext}, \ref{tab:res18ext}, \ref{tab:deittext}, and \ref{tab:swintext}, and Table \ref{tab:vggext2}, \ref{tab:resext2}, \ref{tab:deitext2}, and \ref{tab:swinext2}, respectively. From the $P$ values of paired-sample t-tests, we observed that anatomical structures of the optic cup, optic disk, and blood vessels occupied statistically significant percentages in the focus regions of various glaucoma classifiers. Further, Table \ref{tab:correlation} summarizes Pearson's and Spearman's correlation tests between the model classification performance of AUROC and the anatomies' proportion in the focus area. Notably, there exists a statistically significantly positive relationship between model predictive performance and their focus on anatomical structures, which aligns with the clinician's diagnostic logic of glaucoma \cite{liao2019clinical}.

\begin{figure}
    \centering
    \includegraphics[width=1\linewidth]{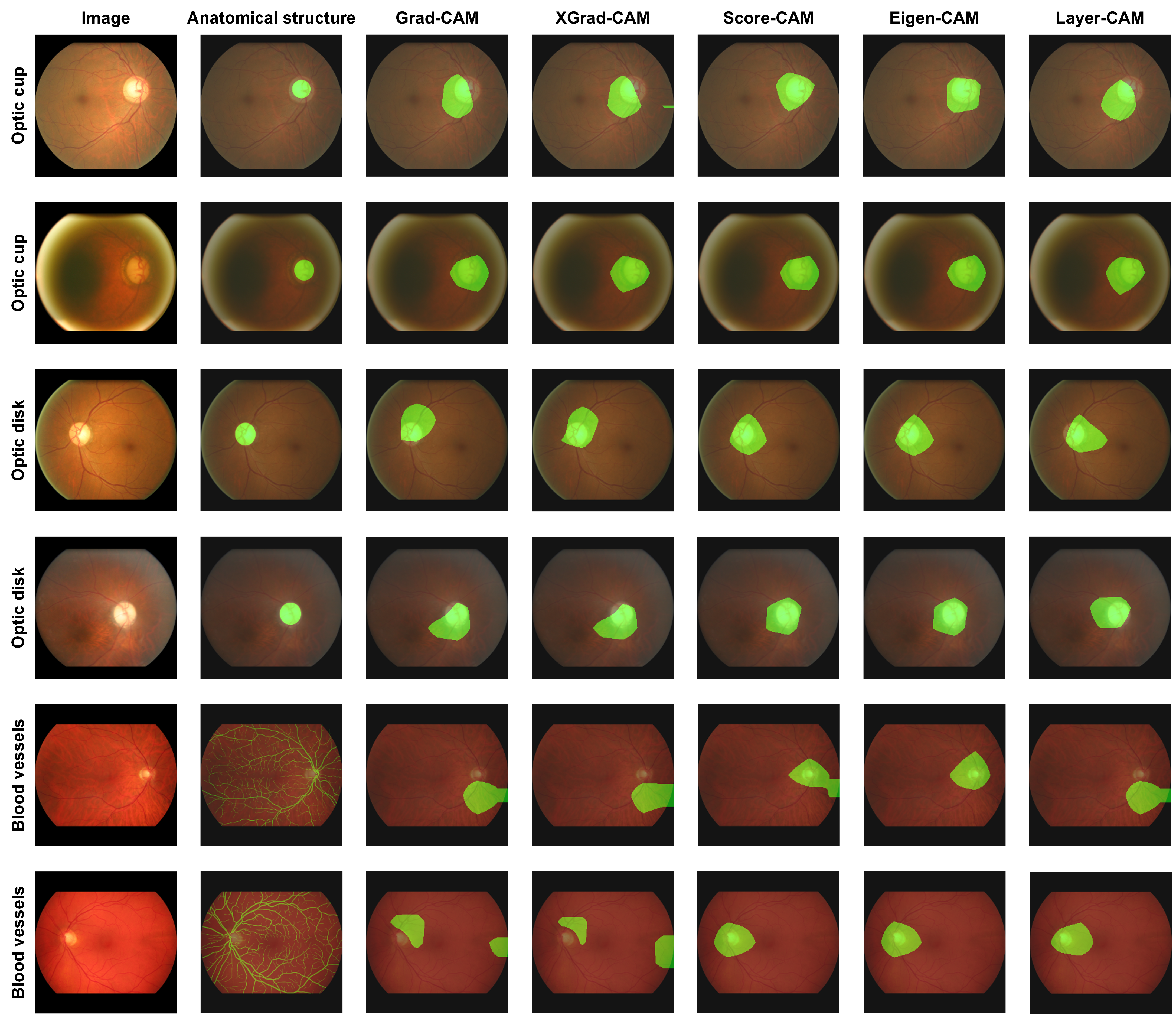}
    \caption{Visualization comparison of anatomical structures and VGG-11 explanations by different CAM methods on the internal test dataset}
    \label{fig1}
\end{figure}

\begin{figure}
    \centering
    \includegraphics[width=1\linewidth]{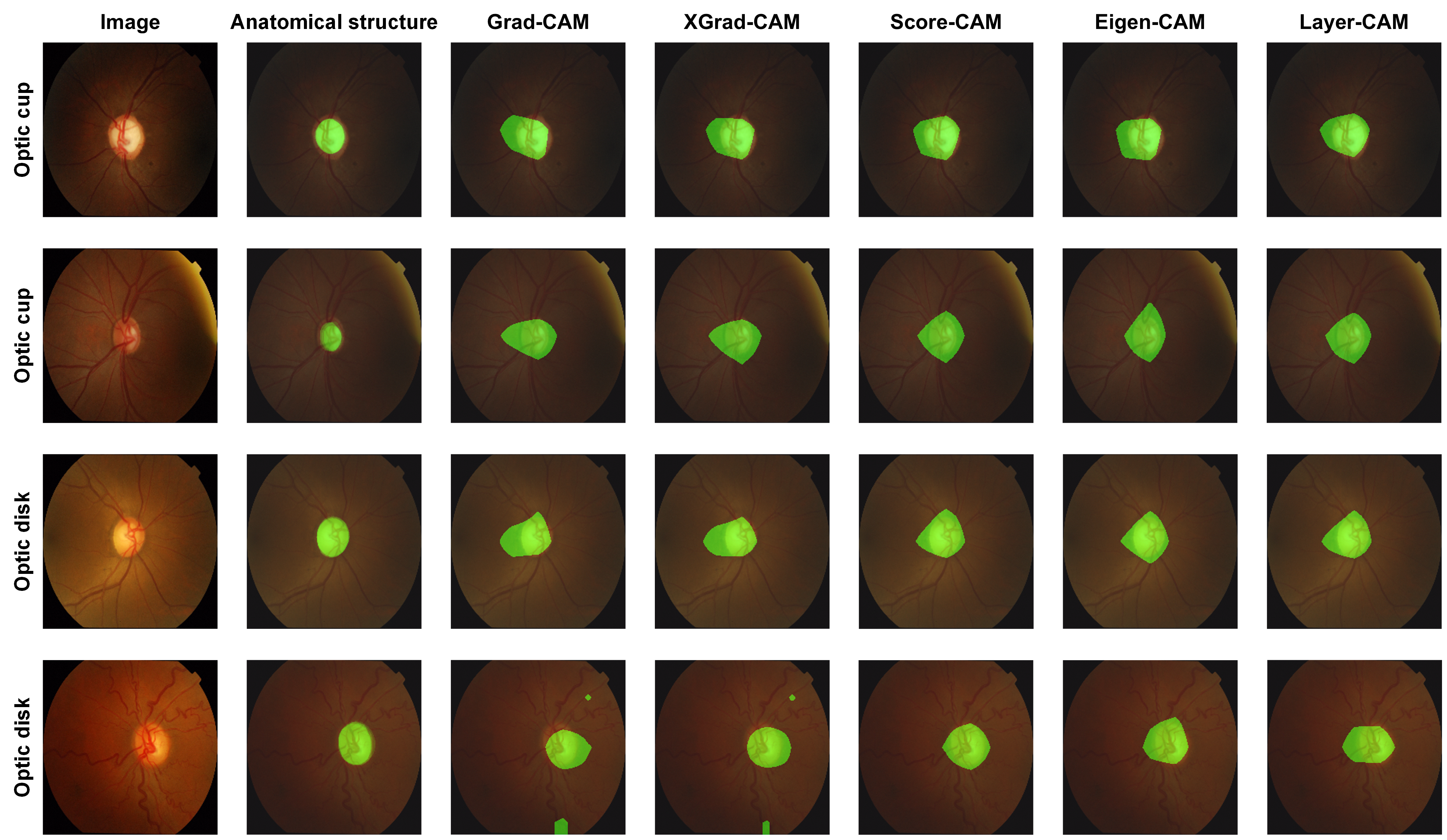}
    \caption{Visualization comparison of anatomical structures and VGG-11 explanations by different CAM methods on the external test dataset of Drishti-GS}
    \label{fig2}
\end{figure}

\begin{figure}
    \centering
    \includegraphics[width=1\linewidth]{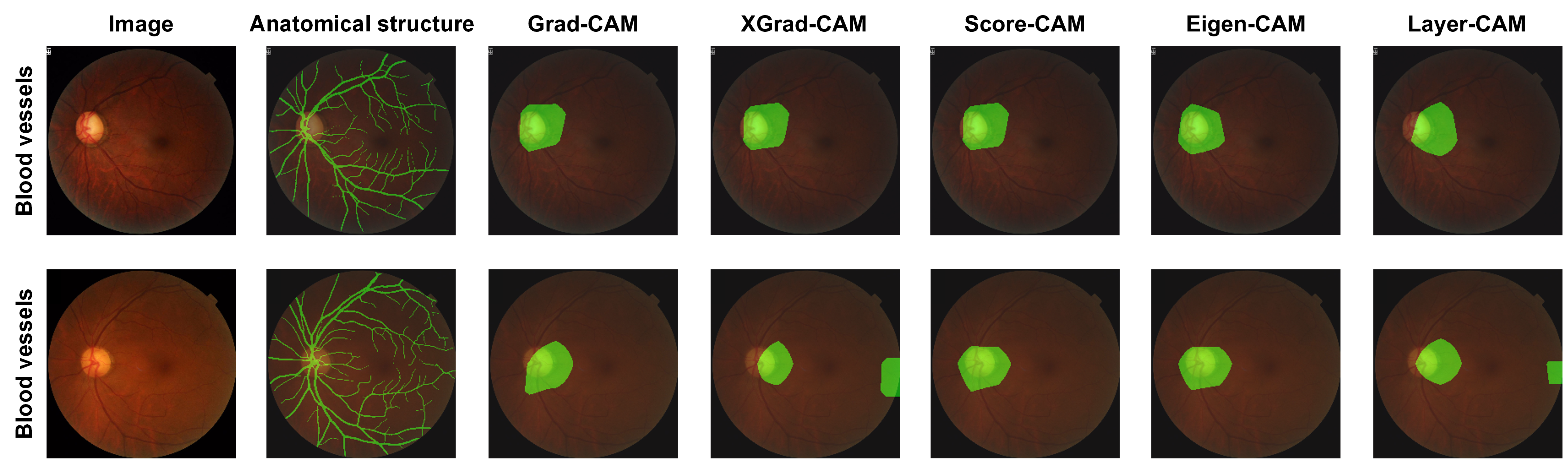}
    \caption{Visualization comparison of anatomical structures and VGG-11 explanations by different CAM methods on the external test dataset of FIVES}
    \label{fig3}
\end{figure}

To provide a comparative visualization of anatomies and model decision logic, Figure \ref{fig1} portrays a comparative visualization of anatomies and VGG-11 decision logic, and Figure \ref{fig2} and \ref{fig3} shows the corresponding results in the external test sets. The images from the first to the fifth column are original images, ground-truth anatomies, and focus area by Grad-CAM, XGrad-CAM, Score-CAM, Eigen-CAM, and Layer-CAM, respectively. The striking resemblance between the model's focus areas and the anatomical regions underscores a compelling point: despite being trained with solely image-level labels, deep neural networks rely on reasonable evidence to classify glaucoma, akin to the decision-making process of human clinicians \cite{liao2019clinical}. This observation also demonstrated the practicability of knowledge distillation by data-driven deep neural networks \cite{knowledge-distillation,2023leveraging,2024clinical}.

\section{Discussion}
In this study, we developed four deep neural networks based on both CNN and Vision Transformer architectures for glaucoma classification. We compared model focus regions generated by different CAM methods and anatomical structures annotated by clinicians to shed light on the extent of alignment between the decision-making logic of black-box deep learning models and the clinical domain knowledge that human experts exploit to make diagnoses. Based on the paired-sample t-test, we showed that the data-driven deep neural networks consistently paid more attention to the anatomical structures of the optic cup, optic disk, and blood vessels. This empirical evidence underscores the convergence of decision logic between our models and human experts in the context of glaucoma classification. Further, we implemented Pearson's and Spearman's correlation analysis and revealed the positive relationship between the model's attentiveness to anatomical structures in the focus area and the model's predictive performance.

In our experiments, VGG-11 outperformed both ResNet-18 and Vision Transformer-based DeiT-Tiny and Swin-Tiny with more sophisticated structure design. Such a phenomenon of "\textit{The deeper is not the better}" has been reported in eye cancer classification \cite{santos2022towards} and pneumonia detection \cite{ikechukwu2021resnet} and is verified by our experimental results on glaucoma classification. We demonstrated that the conventional CNN model of VGG-11 is still capable even achieving the best performance in handling datasets with middle scale and resolution. In terms of AUROC and AUPRC, VGG-11 was 17.9\% and 21.0\% higher than ResNet-18, 9.2\% and 15.3\% higher than DeiT-Tiny, and 9.2\% and 16.6\% higher than Swin-Tiny in the internal test set and a similar substantial superiority can be observed in the external tests sets. Additionally, VGG-11 holds relatively lower complexity than Vision Transformers. The inherent preservation of spatial information in VGG-11 leads to its focus region by various CAM methods overlaps more with the anatomies than the Vision Transformers, making it better aligned with clinical knowledge and becoming more trustworthy from the perspective of healthcare professionals \cite{guide,2024hitl}.

To uncover the model inference logic, we chose various CAM methods to pinpoint the focus regions toward model final decisions. This approach has been employed by prior researchers to segment pixel-level lesion areas based on image-level labels \cite{guide,zhang2020survey,zhang2021weakly,chan2021comprehensive}. However, most of these studies assessed whether the model focused on ROI by calculating overlap metrics such as Intersection over Union (IoU). In this study, we introduced another approach that conducts statistical tests on whether there is a significant difference between the percentage of ROI in the model focus area and the percentage of ROI in the entire image input. Therefore, if the first term is statistically significantly higher than the second term, it is ascertained that the model exhibits a preference for ROI over clinically irrelevant regions. In our case, we showed that the convergence of decision logic in various deep learning models and human clinicians \cite{2021multi}, which was further supported by the positive relationship between model predictive performance and the percentage of anatomies in the model focus area.

Based on this work, there are several limitations to be addressed in the future. First, the explanation methods utilized in this study were confined to the CAM family. Subsequent research should consider the incorporation of additional deep learning explanation methods such as Integrated Gradients \cite{IG} to offer a more comprehensive analysis. Second, this work primarily focuses on explaining the existing fundus images. In future research, we intend to explore an alternative approach that uses image generation methods to manipulate anatomical structures \cite{white2023contrastive,wu2022vessel} and agentic language models to explicitly output decision logic by human language \cite{llms,2025agent}. This approach will allow us to gain deeper insights into the behavior and facilitate users' comprehension of the underlying decision logic of deep neural networks. Third, the statistical tests compared model focus area with anatomies and future work will consider additional clinical evidence such as bleeding and notch \cite{liao2019clinical}. Furthermore, our observations revealed a positive correlation between a model's predictive performance and the percentage of its focus region over anatomical structures. This observation suggests a potential avenue for enhancing model performance, wherein the quantitative percentage of anatomies could be utilized as the rewards under the framework of reinforcement learning \cite{ellis2020impact,kang2}. Finally, this research exclusively explored glaucoma diagnosis and we plan to evaluate the association between model focus area and anatomical structure in a broader spectrum of ophthalmological tasks such as diabetic retinopathy, retinal vein occlusion, and fundus tumors \cite{cen2021automatic}.

\section{Conclusion}
The black-box nature of deep learning has long hindered its application in healthcare. In this study, we validated the statistically significant alignment of clinical domain knowledge and the decision-making logic of deep neural networks through several rigorous statistical tests. We hope that this research mitigates domain experts' concerns regarding the trustworthiness of deep learning in glaucoma diagnosis.

\begin{landscape}
\begin{table}[]
\centering
\caption{An overview of the used dataset. ORIGA, HRF, and LES-AV were used for model development and internal evaluation. Drishiti-GS and FIVES were introduced for external evaluation.}
\label{tab:dataset-overview}
\begin{tabular}{ccccc}
\hline
Dataset    & Function                                                    & Anatomical annotation & Case number & Control number \\ \hline
ORIGA      & \multirow{3}{*}{Model development \& Internal   evaluation} & Optic cup \& disk     & 168         & 482            \\
HRF        &                                                             & Blood vessels         & 15          & 15             \\
LES-AV     &                                                             & Blood vessels         & 11          & 11             \\
Drishti-GS & \multirow{2}{*}{External evaluation}                        & Optic cup \& disk     & 70          & 31             \\
FIVES      &                                                             & Blood vessels         & 150         & 150           \\ \hline
\end{tabular}
\end{table}

\begin{table}[]
\centering
\caption{Data split details for double cross-validation. The split index of (1, 2, 3) denotes a scenario where subset 1 is the training dataset, subset 2 is the validation dataset, and subset 3 is the test dataset. Therefore, we have 6 different scenarios for a comprehensive evaluation of deep learning classification and explanation.}
\label{tab:data-split}
\begin{tabular}{cccc}
\hline
Data split index   & Anatomical annotation & Case number & Control number \\ \hline
\multirow{2}{*}{1} & Optic cup \& disk     & 56          & 161            \\
                   & Blood vessels         & 9           & 9              \\
\multirow{2}{*}{2} & Optic cup \& disk     & 56          & 161            \\
                   & Blood vessels         & 9           & 9              \\
\multirow{2}{*}{3} & Optic cup \& disk     & 56          & 160            \\
                   & Blood vessels         & 8           & 8             \\ \hline
\end{tabular}
\end{table}

\end{landscape}

% \usepackage{lscape}
\begin{landscape}
\begin{table}[]
\centering
\caption{Glaucoma classification performance of various deep learning models on the internal test dataset.}
\label{tab:classification-int}
\begin{tabular}{cccccccc}
\hline
Model     & AUROC           & AUPRC           & Accuracy        & Sensitivity     & Specificity     & PPV             & NPV             \\ \hline
VGG-11    & 0.746   (0.033) & 0.524   (0.062) & 0.699   (0.029) & 0.657   (0.058) & 0.715   (0.034) & 0.471   (0.050) & 0.848   (0.026) \\
ResNet-18 & 0.633   (0.040) & 0.414   (0.054) & 0.622   (0.030) & 0.526   (0.058) & 0.659   (0.035) & 0.374   (0.049) & 0.786   (0.031) \\
DeiT-Tiny & 0.683   (0.037) & 0.444   (0.055) & 0.606   (0.031) & 0.630   (0.054) & 0.597   (0.035) & 0.376   (0.045) & 0.816   (0.030) \\
Swin-Tiny & 0.683   (0.039) & 0.437   (0.057) & 0.623   (0.030) & 0.678   (0.055) & 0.602   (0.036) & 0.396   (0.045) & 0.833   (0.031)
\\ \hline
\end{tabular}
\end{table}

\begin{table}[]
\centering
\caption{Glaucoma classification performance of various deep learning models on the external test dataset of Drishti-GS.}
\label{tab:classification-drishiti}
\begin{tabular}{cccccccc}
\hline
Model     & AUROC           & AUPRC           & Accuracy        & Sensitivity     & Specificity     & PPV             & NPV             \\ \hline
VGG-11    & 0.789   (0.068) & 0.890   (0.048) & 0.686   (0.060) & 0.629   (0.081) & 0.812   (0.091) & 0.881   (0.063) & 0.501   (0.094) \\
ResNet-18 & 0.687   (0.081) & 0.808   (0.073) & 0.526   (0.066) & 0.419   (0.076) & 0.760   (0.095) & 0.793   (0.099) & 0.385   (0.083) \\
DeiT-Tiny & 0.697   (0.075) & 0.841   (0.061) & 0.611   (0.067) & 0.595   (0.076) & 0.646   (0.105) & 0.801   (0.079) & 0.440   (0.107) \\
Swin-Tiny & 0.784   (0.072) & 0.875   (0.054) & 0.673   (0.059) & 0.657   (0.074) & 0.708   (0.105) & 0.839   (0.068) & 0.497   (0.100) \\ \hline
\end{tabular}%
\end{table}

\begin{table}[]
\centering
\caption{Glaucoma classification performance of various deep learning models on the external test dataset of FIVES.}
\label{tab:classification-fives}
\begin{tabular}{cccccccc}
\hline
Model     & AUROC           & AUPRC           & Accuracy        & Sensitivity     & Specificity     & PPV             & NPV             \\ \hline
VGG-11    & 0.733   (0.040) & 0.771   (0.036) & 0.691   (0.039) & 0.596   (0.055) & 0.787   (0.044) & 0.743   (0.051) & 0.664   (0.048) \\
ResNet-18 & 0.661   (0.044) & 0.662   (0.052) & 0.605   (0.038) & 0.571   (0.052) & 0.638   (0.051) & 0.614   (0.054) & 0.599   (0.049) \\
DeiT-Tiny & 0.629   (0.045) & 0.644   (0.051) & 0.586   (0.039) & 0.484   (0.051) & 0.686   (0.046) & 0.627   (0.058) & 0.576   (0.048) \\
Swin-Tiny & 0.570   (0.046) & 0.611   (0.052) & 0.566   (0.042) & 0.482   (0.058) & 0.649   (0.051) & 0.583   (0.058) & 0.556   (0.051) \\ \hline
\end{tabular}%
\end{table}

\end{landscape}

% \usepackage{multirow}
% \usepackage{graphicx}
% \usepackage{lscape}
\begin{landscape}
\begin{table}[]
\centering
\caption{Glaucoma explanation performance of various XAI methods integrated with VGG-11 on the internal test dataset. Top 5\% activated pixels were selected as the model focus area.}
\label{tab:VGG-11}
\begin{tabular}{ccccccc}
\hline
Model                    & XAI                        & Anatomy       & Activation ratio (\%) & Structure ratio (\%) & Ratio difference (\%) & P value   \\ \hline
\multirow{15}{*}{VGG-11} & \multirow{3}{*}{Grad-CAM}  & Optic cup     & 4.74 (0.18)           & 0.71 (0.00)          & 4.03 (0.18)           & 4.19E-77  \\
                         &                            & Optic disk    & 10.36 (0.36)          & 1.80 (0.00)          & 8.56 (0.36)           & 9.12E-85  \\
                         &                            & Blood vessels & 22.95 (0.94)          & 18.96 (0.28)         & 3.98 (0.84)           & 1.66E-04  \\
                         & \multirow{3}{*}{XGrad-CAM} & Optic cup     & 5.68 (0.20)           & 0.71 (0.00)          & 4.97 (0.18)           & 4.11E-100 \\
                         &                            & Optic disk    & 12.33 (0.41)          & 1.80 (0.00)          & 10.53 (0.41)          & 5.63E-112 \\
                         &                            & Blood vessels & 23.15 (0.92)          & 18.96 (0.28)         & 4.18 (1.05)           & 1.67E-04  \\
                         & \multirow{3}{*}{Score-CAM} & Optic cup     & 10.22 (0.18)          & 0.71 (0.00)          & 9.51 (0.18)           & 3.48E-259 \\
                         &                            & Optic disk    & 24.07 (0.38)          & 1.80 (0.00)          & 22.27 (0.38)          & 2.26E-304 \\
                         &                            & Blood vessels & 33.23 (0.94)          & 18.96 (0.28)         & 14.26 (0.92)          & 1.93E-25  \\
                         & \multirow{3}{*}{Eigen-CAM} & Optic cup     & 11.92 (0.18)          & 0.71 (0.00)          & 11.21 (0.18)          & 0.00E+00  \\
                         &                            & Optic disk    & 29.21 (0.33)          & 1.80 (0.00)          & 27.41 (0.33)          & 0.00E+00  \\
                         &                            & Blood vessels & 41.33 (0.92)          & 18.96 (0.28)         & 22.37 (0.84)          & 8.39E-48  \\
                         & \multirow{3}{*}{Layer-CAM} & Optic cup     & 12.34 (0.18)          & 0.71 (0.00)          & 11.63 (0.18)          & 0.00E+00  \\
                         &                            & Optic disk    & 29.20 (0.31)          & 1.80 (0.00)          & 27.40 (0.31)          & 0.00E+00  \\
                         &                            & Blood vessels & 33.66 (0.94)          & 18.96 (0.28)         & 14.70 (1.07)          & 1.61E-23 \\ \hline
\end{tabular}
\end{table}
\end{landscape}

% \usepackage{multirow}
% \usepackage{graphicx}
% \usepackage{lscape}
\begin{landscape}
\begin{table}[]
\centering
\caption{Glaucoma explanation performance of various XAI methods integrated with ResNet-18 on the internal test dataset. Top 5\% activated pixels were selected as the model focus area.}
\label{tab:ResNet-18}
%\resizebox{\columnwidth}{!}{%
\begin{tabular}{ccccccc}
\hline
\multicolumn{1}{c}{Model}   & \multicolumn{1}{c}{XAI}    & \multicolumn{1}{c}{Anatomy} & \multicolumn{1}{c}{Activation ratio (\%)} & \multicolumn{1}{c}{Structure ratio (\%)} & \multicolumn{1}{c}{Ratio difference (\%)} & \multicolumn{1}{c}{P value} \\ \hline
\multirow{15}{*}{ResNet-18} & \multirow{3}{*}{Grad-CAM}  & Optic cup                   & 3.28   (0.18)                             & 0.71   (0.00)                            & 2.57   (0.18)                             & 2.50E-46                    \\
                            &                            & Optic disk                  & 6.97   (0.36)                             & 1.80   (0.00)                            & 5.17   (0.33)                             & 3.41E-48                    \\
                            &                            & Blood vessels               & 22.43   (1.68)                            & 18.96   (0.28)                           & 3.47   (1.73)                             & 4.83E-02                    \\
                            & \multirow{3}{*}{XGrad-CAM} & Optic cup                   & 1.07   (0.08)                             & 0.71   (0.00)                            & 0.36   (0.08)                             & 1.48E-04                    \\
                            &                            & Optic disk                  & 2.56   (0.13)                             & 1.80   (0.00)                            & 0.76   (0.13)                             & 1.06E-04                    \\
                            &                            & Blood vessels               & 23.02   (1.25)                            & 18.96   (0.28)                           & 4.06   (1.20)                             & 1.23E-03                    \\
                            & \multirow{3}{*}{Score-CAM} & Optic cup                   & 2.89   (0.18)                             & 0.71   (0.00)                            & 2.17   (0.18)                             & 4.39E-41                    \\
                            &                            & Optic disk                  & 6.63   (0.38)                             & 1.80   (0.00)                            & 4.83   (0.38)                             & 4.35E-45                    \\
                            &                            & Blood vessels               & 31.08   (1.12)                            & 18.96   (0.28)                           & 12.12   (1.10)                            & 4.31E-20                    \\
                            & \multirow{3}{*}{Eigen-CAM} & Optic cup                   & 4.68   (0.18)                             & 0.71   (0.00)                            & 3.96   (0.18)                             & 7.22E-86                    \\
                            &                            & Optic disk                  & 10.75   (0.36)                            & 1.80   (0.00)                            & 8.95   (0.36)                             & 6.72E-104                   \\
                            &                            & Blood vessels               & 33.88   (0.84)                            & 18.96   (0.28)                           & 14.92   (0.87)                            & 7.76E-32                    \\
                            & \multirow{3}{*}{Layer-CAM} & Optic cup                   & 5.66   (0.20)                             & 0.71   (0.00)                            & 4.95   (0.18)                             & 8.76E-128                   \\
                            &                            & Optic disk                  & 12.89   (0.38)                            & 1.80   (0.00)                            & 11.10   (0.38)                            & 5.64E-154                   \\
                            &                            & Blood vessels               & 34.41   (0.66)                            & 18.96   (0.28)                           & 15.45   (0.69)                            & 1.26E-35                   \\ \hline
\end{tabular}
%}
\end{table}
\end{landscape}

% \usepackage{multirow}
% \usepackage{graphicx}
% \usepackage{lscape}
\begin{landscape}
\begin{table}[]
\centering
\caption{Glaucoma explanation performance of various XAI methods integrated with DeiT-Tiny on the internal test dataset. Top 5\% activated pixels were selected as the model focus area.}
\label{tab:DeiT-Tiny}
%\resizebox{\columnwidth}{!}{%
\begin{tabular}{ccccccc}
\hline
\multicolumn{1}{c}{Model}   & \multicolumn{1}{c}{XAI}    & \multicolumn{1}{c}{Anatomy} & \multicolumn{1}{c}{Activation ratio (\%)} & \multicolumn{1}{c}{Structure ratio (\%)} & \multicolumn{1}{c}{Ratio difference (\%)} & \multicolumn{1}{c}{P value} \\ \hline
\multirow{15}{*}{DeiT-Tiny} & \multirow{3}{*}{Grad-CAM}  & Optic cup                   & 3.52   (0.13)                             & 0.71   (0.00)                            & 2.81   (0.15)                             & 4.90E-53                    \\
                            &                            & Optic disk                  & 7.25   (0.28)                             & 1.80   (0.00)                            & 5.45   (0.28)                             & 4.38E-52                    \\
                            &                            & Blood vessels               & 22.57   (1.28)                            & 18.96   (0.28)                           & 3.61   (1.20)                             & 3.63E-03                    \\
                            & \multirow{3}{*}{XGrad-CAM} & Optic cup                   & 3.51   (0.15)                             & 0.71   (0.00)                            & 2.79   (0.15)                             & 5.84E-63                    \\
                            &                            & Optic disk                  & 7.99   (0.31)                             & 1.80   (0.00)                            & 6.19   (0.31)                             & 7.93E-69                    \\
                            &                            & Blood vessels               & 19.09   (1.15)                            & 18.96   (0.28)                           & 0.13   (1.12)                             & 9.12E-01                    \\
                            & \multirow{3}{*}{Score-CAM} & Optic cup                   & 2.84   (0.18)                             & 0.71   (0.00)                            & 2.13   (0.18)                             & 1.69E-34                    \\
                            &                            & Optic disk                  & 5.76   (0.36)                             & 1.80   (0.00)                            & 3.96   (0.36)                             & 2.07E-31                    \\
                            &                            & Blood vessels               & 19.71   (1.28)                            & 18.96   (0.28)                           & 0.75   (1.20)                             & 5.42E-01                    \\
                            & \multirow{3}{*}{Eigen-CAM} & Optic cup                   & 1.81   (0.13)                             & 0.71   (0.00)                            & 1.10   (0.13)                             & 1.90E-16                    \\
                            &                            & Optic disk                  & 4.31   (0.23)                             & 1.80   (0.00)                            & 2.51   (0.23)                             & 8.91E-18                    \\
                            &                            & Blood vessels               & 24.67   (0.97)                            & 18.96   (0.28)                           & 5.70   (0.97)                             & 1.76E-07                    \\
                            & \multirow{3}{*}{Layer-CAM} & Optic cup                   & 7.43   (0.20)                             & 0.71   (0.00)                            & 6.72   (0.20)                             & 7.85E-174                   \\
                            &                            & Optic disk                  & 16.18   (0.38)                            & 1.80   (0.00)                            & 14.38   (0.38)                            & 4.41E-195                   \\
                            &                            & Blood vessels               & 25.32   (1.02)                            & 18.96   (0.28)                           & 6.36   (1.10)                             & 6.46E-07                   \\ \hline
\end{tabular}
\end{table}
\end{landscape}

% \usepackage{multirow}
% \usepackage{graphicx}
% \usepackage{lscape}
\begin{landscape}
\begin{table}[]
\centering
\caption{Glaucoma explanation performance of various XAI methods integrated with Swin-Tiny on the internal test dataset. Top 5\% activated pixels were selected as the model focus area.}
\label{tab:Swin-Tiny}
%\resizebox{\columnwidth}{!}{%
\begin{tabular}{ccccccc}
\hline
Model                       & XAI                        & Anatomy       & Activation ratio (\%) & Structure ratio (\%) & Ratio difference (\%) & P value   \\ \hline
\multirow{15}{*}{Swin-Tiny} & \multirow{3}{*}{Grad-CAM}  & Optic cup     & 5.10   (0.20)         & 0.71   (0.00)        & 4.39   (0.20)         & 4.97E-77  \\
                            &                            & Optic disk    & 11.33   (0.46)        & 1.80   (0.00)        & 9.53   (0.46)         & 5.91E-84  \\
                            &                            & Blood vessels & 31.42   (1.53)        & 18.96   (0.28)       & 12.46   (1.38)        & 1.98E-15  \\
                            & \multirow{3}{*}{XGrad-CAM} & Optic cup     & 2.90   (0.15)         & 0.71   (0.00)        & 2.18   (0.15)         & 8.19E-38  \\
                            &                            & Optic disk    & 6.68   (0.36)         & 1.80   (0.00)        & 4.88 (0.36)           & 1.03E-41  \\
                            &                            & Blood vessels & 25.52   (1.28)        & 18.96   (0.28)       & 6.56   (1.25)         & 7.32E-07  \\
                            & \multirow{3}{*}{Score-CAM} & Optic cup     & 9.24   (0.23)         & 0.71   (0.00)        & 8.53   (0.20)         & 3.47E-204 \\
                            &                            & Optic disk    & 21.65   (0.46)        & 1.80   (0.00)        & 19.85   (0.46)        & 1.05E-233 \\
                            &                            & Blood vessels & 32.84   (1.35)        & 18.96   (0.28)       & 13.88   (1.38)        & 1.52E-15  \\
                            & \multirow{3}{*}{Eigen-CAM} & Optic cup     & 6.27   (0.18)         & 0.71   (0.00)        & 5.56   (0.18)         & 1.52E-125 \\
                            &                            & Optic disk    & 13.53   (0.36)        & 1.80   (0.00)        & 11.73   (0.36)        & 2.49E-130 \\
                            &                            & Blood vessels & 35.76   (1.43)        & 18.96   (0.28)       & 16.80   (1.35)        & 2.14E-24  \\
                            & \multirow{3}{*}{Layer-CAM} & Optic cup     & 6.88   (0.18)         & 0.71   (0.00)        & 6.17   (0.18)         & 2.21E-145 \\
                            &                            & Optic disk    & 15.79   (0.38)        & 1.80   (0.00)        & 13.99   (0.38)        & 6.39E-164 \\
                            &                            & Blood vessels & 30.07   (1.07)        & 18.96   (0.28)       & 11.11   (1.15)        & 2.95E-15 \\ \hline
\end{tabular}
\end{table}
\end{landscape}

\begin{landscape}
\begin{table}[]
\centering
\caption{Glaucoma explanation performance of various XAI methods integrated with VGG-11 on the external test dataset of Drishti-GS. Top 5\% activated pixels were selected as the model focus area.}
\label{tab:vgg11ext}
\begin{tabular}{ccccccc}
\hline
Model                    & XAI                        & Anatomy    & Activation ratio (\%) & Structure ratio (\%) & Ratio difference (\%) & P value   \\ \hline
\multirow{10}{*}{VGG-11} & \multirow{2}{*}{Grad-CAM}  & Optic cup  & 10.23   (0.84)        & 2.02   (0.05)        & 8.22   (0.84)         & 5.53E-18  \\
                         &                            & Optic disk & 14.57   (1.17)        & 3.44   (0.03)        & 11.13   (1.20)        & 6.77E-18  \\
                         & \multirow{2}{*}{XGrad-CAM} & Optic cup  & 13.07   (0.82)        & 2.02   (0.05)        & 11.05   (0.79)        & 7.59E-26  \\
                         &                            & Optic disk & 18.64   (1.12)        & 3.44   (0.03)        & 15.19   (1.10)        & 1.55E-26  \\
                         & \multirow{2}{*}{Score-CAM} & Optic cup  & 27.72   (0.99)        & 2.02   (0.05)        & 25.70   (0.94)        & 4.14E-65  \\
                         &                            & Optic disk & 41.05   (1.43)        & 3.44   (0.03)        & 37.60   (1.43)        & 6.44E-71  \\
                         & \multirow{2}{*}{Eigen-CAM} & Optic cup  & 33.12   (0.84)        & 2.02   (0.05)        & 31.10   (0.82)        & 2.53E-97  \\
                         &                            & Optic disk & 52.27   (1.17)        & 3.44   (0.03)        & 48.83   (1.17)        & 1.34E-125 \\
                         & \multirow{2}{*}{Layer-CAM} & Optic cup  & 34.60   (0.87)        & 2.02   (0.05)        & 32.59   (0.84)        & 1.66E-105 \\
                         &                            & Optic disk & 52.58   (1.10)        & 3.44   (0.03)        & 49.13   (1.10)        & 5.34E-127 \\ \hline
\end{tabular}
\end{table}

\begin{table}[]
\centering
\caption{Glaucoma explanation performance of various XAI methods integrated with ResNet-18 on the external test dataset od Drishti-GS. Top 5\% activated pixels were selected as the model focus area.}
\label{tab:res18ext}
\begin{tabular}{ccccccc}
\hline
Model                      & XAI                        & Anatomy    & Activation ratio (\%) & Structure ratio (\%) & Ratio difference (\%) & P value   \\ \hline
\multirow{9}{*}{ResNet-18} & \multirow{2}{*}{Grad-CAM}  & Optic cup  & 15.78   (1.17)        & 2.02   (0.05)        & 13.77   (1.15)        & 4.55E-28  \\
                           &                            & Optic disk & 22.05   (1.53)        & 3.44   (0.03)        & 18.60   (1.53)        & 4.79E-28  \\
                           & \multirow{2}{*}{XGrad-CAM} & Optic cup  & 5.43   (0.61)         & 2.02   (0.05)        & 3.41   (0.59)         & 4.40E-07  \\
                           &                            & Optic disk & 8.17   (0.84)         & 3.44   (0.03)        & 4.73   (0.84)         & 4.63E-07  \\
                           & \multirow{2}{*}{Score-CAM} & Optic cup  & 16.25   (0.92)        & 2.02   (0.05)        & 14.23   (0.92)        & 6.15E-33  \\
                           &                            & Optic disk & 23.50   (1.28)        & 3.44   (0.03)        & 20.06   (1.28)        & 8.22E-34  \\
                           & \multirow{2}{*}{Eigen-CAM} & Optic cup  & 20.34   (1.02)        & 2.02   (0.05)        & 18.32   (1.02)        & 4.05E-41  \\
                           &                            & Optic disk & 30.52   (1.51)        & 3.44   (0.03)        & 27.08   (1.51)        & 3.19E-45  \\
                           & \multirow{2}{*}{Layer-CAM} & Optic cup  & 17.18   (1.12)        & 2.02   (0.05)        & 15.17   (1.12)        & 4.56E-35  \\
                           &                            & Optic disk & 52.58   (1.10)        & 3.44   (0.03)        & 49.13   (1.10)        & 5.34E-127 \\ \hline
\end{tabular}
\end{table}
\end{landscape}

% \usepackage{multirow}
% \usepackage{lscape}
\begin{landscape}
\begin{table}[]
\centering
\caption{Glaucoma explanation performance of various XAI methods integrated with DeiT-Tiny on the external test dataset of Drishti-GS. Top 5\% activated pixels were selected as the model focus area.}
\label{tab:deittext}
\begin{tabular}{ccccccc}
\hline
Model                       & XAI                        & Anatomy    & Activation ratio (\%) & Structure ratio (\%) & Ratio difference (\%) & P value  \\ \hline
\multirow{10}{*}{DeiT-Tiny} & \multirow{2}{*}{Grad-CAM}  & Optic cup  & 11.95   (0.87)        & 2.02   (0.05)        & 9.94   (0.87)         & 2.35E-20 \\
                            &                            & Optic disk & 17.58   (1.28)        & 3.44   (0.03)        & 14.13   (1.28)        & 2.21E-20 \\
                            & \multirow{2}{*}{XGrad-CAM} & Optic cup  & 10.42   (0.79)        & 2.02   (0.05)        & 8.41   (0.79)         & 1.29E-19 \\
                            &                            & Optic disk & 15.09   (1.10)        & 3.44   (0.03)        & 11.65   (1.10)        & 7.11E-20 \\
                            & \multirow{2}{*}{Score-CAM} & Optic cup  & 7.85   (0.82)         & 2.02   (0.05)        & 5.83   (0.77)         & 1.35E-10 \\
                            &                            & Optic disk & 11.73   (1.10)        & 3.44   (0.03)        & 8.29   (1.10)         & 2.82E-10 \\
                            & \multirow{2}{*}{Eigen-CAM} & Optic cup  & 6.39   (0.87)         & 2.02   (0.05)        & 4.38   (0.87)         & 1.25E-07 \\
                            &                            & Optic disk & 10.48   (1.17)        & 3.44   (0.03)        & 7.04   (1.17)         & 6.94E-09 \\
                            & \multirow{2}{*}{Layer-CAM} & Optic cup  & 19.71   (0.69)        & 2.02   (0.05)        & 17.69   (0.69)        & 4.33E-58 \\
                            &                            & Optic disk & 28.62   (0.89)        & 3.44   (0.03)        & 25.17   (0.92)        & 1.01E-64 \\ \hline
\end{tabular}
\end{table}

\begin{table}[]
\centering
\caption{Glaucoma explanation performance of various XAI methods integrated with Swin-Tiny on the external test dataset of Drishti-GS. Top 5\% activated pixels were selected as the model focus area.}
\label{tab:swintext}
\begin{tabular}{ccccccc}
\hline
Model                       & XAI                        & Anatomy    & Activation ratio (\%) & Structure ratio (\%) & Ratio difference (\%) & P value  \\ \hline
\multirow{10}{*}{Swin-Tiny} & \multirow{2}{*}{Grad-CAM}  & Optic cup  & 19.66   (1.02)        & 2.02   (0.05)        & 17.65   (0.99)        & 3.03E-36 \\
                            &                            & Optic disk & 28.57   (1.43)        & 3.44   (0.03)        & 25.12   (1.45)        & 3.36E-37 \\
                            & \multirow{2}{*}{XGrad-CAM} & Optic cup  & 12.97   (0.99)        & 2.02   (0.05)        & 10.96   (0.99)        & 2.92E-22 \\
                            &                            & Optic disk & 19.15   (1.33)        & 3.44   (0.03)        & 15.71   (1.33)        & 1.85E-23 \\
                            & \multirow{2}{*}{Score-CAM} & Optic cup  & 28.56   (1.05)        & 2.02   (0.05)        & 26.55   (1.02)        & 1.30E-66 \\
                            &                            & Optic disk & 42.79   (1.58)        & 3.44   (0.03)        & 39.34   (1.58)        & 1.07E-72 \\
                            & \multirow{2}{*}{Eigen-CAM} & Optic cup  & 25.04   (1.15)        & 2.02   (0.05)        & 23.02   (1.15)        & 5.47E-52 \\
                            &                            & Optic disk & 36.57   (1.61)        & 3.44   (0.03)        & 33.12   (1.61)        & 1.30E-54 \\
                            & \multirow{2}{*}{Layer-CAM} & Optic cup  & 17.44   (0.89)        & 2.02   (0.05)        & 15.43   (0.87)        & 5.27E-38 \\
                            &                            & Optic disk & 26.21   (1.15)        & 3.44   (0.03)        & 22.77   (1.12)        & 8.48E-42 \\ \hline
\end{tabular}
\end{table}
\end{landscape}

\begin{landscape}
\begin{table}[]
\centering
\caption{Glaucoma explanation performance of various XAI methods integrated with VGG-11 on the external test dataset of FIVES. Top 5\% activated pixels were selected as the model focus area.}
\label{tab:vggext2}
\begin{tabular}{ccccccc}
\hline
Model                   & XAI       & Anatomy       & Activation ratio (\%) & Structure ratio (\%) & Ratio difference (\%) & P value   \\ \hline
\multirow{5}{*}{VGG-11} & Grad-CAM  & Blood vessels & 17.48   (0.31)        & 14.29   (0.10)       & 3.19   (0.38)         & 3.46E-17  \\
                        & XGrad-CAM & Blood vessels & 17.02   (0.31)        & 14.29   (0.10)       & 2.73   (0.36)         & 1.01E-12  \\
                        & Score-CAM & Blood vessels & 24.11   (0.38)        & 14.29   (0.10)       & 9.82   (0.38)         & 6.02E-93  \\
                        & Eigen-CAM & Blood vessels & 30.67   (0.46)        & 14.29   (0.10)       & 16.38   (0.38)        & 3.39E-218 \\
                        & Layer-CAM & Blood vessels & 25.30   (0.33)        & 14.29   (0.10)       & 11.01   (0.36)        & 2.35E-134 \\ \hline
\end{tabular}
\end{table}

\begin{table}[]
\centering
\caption{Glaucoma explanation performance of various XAI methods integrated with ResNet-18 on the external test dataset of FIVES. Top 5\% activated pixels were selected as the model focus area.}
\label{tab:resext2}
\begin{tabular}{ccccccc}
\hline
Model                      & XAI       & Anatomy       & Activation ratio (\%) & Structure ratio (\%) & Ratio difference (\%) & P value   \\ \hline
\multirow{5}{*}{ResNet-18} & Grad-CAM  & Blood vessels & 19.60   (0.46)        & 14.29   (0.10)       & 5.31   (0.48)         & 1.91E-25  \\
                           & XGrad-CAM & Blood vessels & 16.77   (0.31)        & 14.29   (0.10)       & 2.48   (0.33)         & 8.98E-14  \\
                           & Score-CAM & Blood vessels & 23.59   (0.31)        & 14.29   (0.10)       & 9.30   (0.28)         & 6.16E-116 \\
                           & Eigen-CAM & Blood vessels & 27.89   (0.38)        & 14.29   (0.10)       & 13.60   (0.33)        & 4.44E-196 \\
                           & Layer-CAM & Blood vessels & 26.56   (0.38)        & 14.29   (0.10)       & 12.27   (0.33)        & 3.14E-183 \\ \hline
\end{tabular}
\end{table}

\begin{table}[]
\centering
\caption{Glaucoma explanation performance of various XAI methods integrated with DeiT-Tiny on the external test dataset of FIVES. Top 5\% activated pixels were selected as the model focus area.}
\label{tab:deitext2}
\begin{tabular}{ccccccc}
\hline
Model                      & XAI       & Anatomy       & Activation ratio (\%) & Structure ratio (\%) & Ratio difference (\%) & P value  \\ \hline
\multirow{5}{*}{DeiT-Tiny} & Grad-CAM  & Blood vessels & 16.41   (0.36)        & 14.29   (0.10)       & 2.12   (0.36)         & 1.56E-08 \\
                           & XGrad-CAM & Blood vessels & 15.79   (0.36)        & 14.29   (0.10)       & 1.50   (0.33)         & 3.29E-06 \\
                           & Score-CAM & Blood vessels & 14.97   (0.41)        & 14.29   (0.10)       & 0.68   (0.41)         & 1.05E-01 \\
                           & Eigen-CAM & Blood vessels & 14.88   (0.31)        & 14.29   (0.10)       & 0.59   (0.28)         & 3.05E-02 \\
                           & Layer-CAM & Blood vessels & 20.78   (0.36)        & 14.29   (0.10)       & 6.49   (0.33)         & 1.73E-65 \\ \hline
\end{tabular}
\end{table}

\end{landscape}

\begin{landscape}
\begin{table}[]
\centering
\caption{Glaucoma explanation performance of various XAI methods integrated with Swin-Tiny on the external test dataset of FIVES. Top 5\% activated pixels were selected as the model focus area.}
\label{tab:swinext2}
\begin{tabular}{ccccccc}
\hline
Model                      & XAI       & Anatomy       & Activation ratio (\%) & Structure ratio (\%) & Ratio difference (\%) & P value   \\ \hline
\multirow{5}{*}{Swin-Tiny} & Grad-CAM  & Blood vessels & 21.40   (0.38)        & 14.29   (0.10)       & 7.11   (0.41)         & 8.20E-55  \\
                           & XGrad-CAM & Blood vessels & 18.96   (0.41)        & 14.29   (0.10)       & 4.67   (0.36)         & 3.50E-31  \\
                           & Score-CAM & Blood vessels & 26.53   (0.38)        & 14.29   (0.10)       & 12.24   (0.38)        & 9.14E-128 \\
                           & Eigen-CAM & Blood vessels & 20.61   (0.54)        & 14.29   (0.10)       & 6.32   (0.43)         & 2.87E-42  \\
                           & Layer-CAM & Blood vessels & 21.31   (0.41)        & 14.29   (0.10)       & 7.02   (0.38)         & 2.13E-62  \\ \hline
\end{tabular}
\end{table}

\begin{table}[]
\centering
\caption{Pearson's and Spearman's correlation coefficient between model predictive performance measured by AUROC and model focus area of activation ratio on anatomical structures. \newline}
\label{tab:correlation}
%\resizebox{\columnwidth}{!}{%
\begin{tabular}{cccccc}
\hline
Evaluation                & Anatomy       & Pearson's Correlation & P value  & Spearman's Correlation & P value  \\ \hline
\multirow{3}{*}{Internal} & Optic cup     & 0.42                  & 2.04E-06 & 0.42                   & 1.79E-06 \\
                          & Optic disk    & 0.43                  & 1.26E-06 & 0.41                   & 3.77E-06 \\
                          & Blood vessels & 0.13                  & 1.57E-01 & 0.16                   & 7.81E-02 \\
\multirow{3}{*}{External} & Optic cup     & 0.27                  & 2.51E-03 & 0.29                   & 1.19E-03 \\
                          & Optic disk    & 0.28                  & 1.91E-03 & 0.29                   & 1.29E-03 \\
                          & Blood vessels & 0.27                  & 2.40E-03 & 0.28                   & 1.77E-03 \\ \hline
\end{tabular}
%}
\end{table}
\end{landscape}

\clearpage
\bibliography{Reference}
\end{document}